%% file: acl2021.tex
\documentclass[11pt,a4paper]{article}
\usepackage[hyperref]{acl2021}
\usepackage{times}
\usepackage{latexsym}
\usepackage{amsmath}
\usepackage{tabularx}
\usepackage{mathtools}
\usepackage{makecell}
\usepackage{listliketab}
\usepackage[export]{adjustbox}
\DeclareMathOperator*{\argmax}{argmax}
\usepackage[ruled,vlined]{algorithm2e}
\usepackage[size=small]{todonotes}
\usepackage{booktabs}
\usepackage{enumitem}

\newcommand{\ours}{BERT-Defense}
\newcommand{\basedist}{attack-agnostic}
\usepackage{makecell}
\newcommand{\fulldist}{domain-specific}

\usepackage{microtype}

\aclfinalcopy 
\def\aclpaperid{1234} 

\title{\ours{}: A Probabilistic Model Based on BERT to \\
Combat Cognitively Inspired Orthographic Adversarial Attacks}

\author{Yannik Keller$^\dagger$, Jan Mackensen$^\dagger$, Steffen Eger$\ddagger$ \\
  $^\dagger$ Center of Cognitive Science \\
  $^\ddagger$ Natural Language Learning Group\\
  Technische Universität Darmstadt \\
  \texttt{\{yannik.keller,jan.mackensen\}@stud.tu-darmstadt.de}\\
  \texttt{eger@aiphes.tu-darmstadt.de}
  }

\date{}

\begin{document}
\maketitle
\input{abstract}

\setlength{\abovedisplayskip}{4.5pt}
\setlength{\belowdisplayskip}{4.5pt}

\input{introduction}

\input{related}
\input{method}
\input{experiments}
\input{results}
\input{conclusion}
\section*{Acknowledgments}
We thank the anonymous reviewers for their helpful comments. 
\clearpage
\bibliography{zeroe,acl2021}
\bibliographystyle{acl_natbib}
\clearpage
\input{appendix}

\end{document}

%% file: abstract.tex
\begin{abstract}
Adversarial attacks expose important blind spots of deep learning systems. While word- and sentence-level attack scenarios mostly deal with finding semantic paraphrases of the input that fool NLP models, character-level attacks 
typically insert typos into the input stream. 
It is commonly thought that these are easier to defend via spelling correction modules. In this work, we show that both a standard spellchecker and the approach of \citet{pruthi-etal-2019-combating}, which trains to defend against insertions, deletions and swaps, perform poorly 
on the character-level benchmark recently proposed in \citet{benz} which includes more challenging attacks such as visual and phonetic perturbations and missing word segmentations. In contrast, we show that an untrained iterative approach which combines context-independent character-level information with context-dependent information from BERT's masked language modeling can perform on par with human crowd-workers from Amazon Mechanical Turk (AMT) supervised via 3-shot learning.  \end{abstract}

%% file: introduction.tex
\section{Introduction}

Adversarial attacks to machine learning systems are malicious 
modifications of their inputs designed to fool machines into misclassification but not humans \citep{Goodfellow2014Explaining}. One of their goals 
is to expose blind-spots of deep learning models, which can then be shielded against. In the NLP community, typically two different kinds of attack scenarios are considered. ``High-level'' attacks paraphrase (semantically or syntactically) the input sentence \citep{Iyyer2018Adversarial,Alzantot2018Generating,Jin:2020} so that the classification label does not change, but the model changes its decision.
\begin{figure}[!t]
    \centering
    \includegraphics[width=0.75\columnwidth]{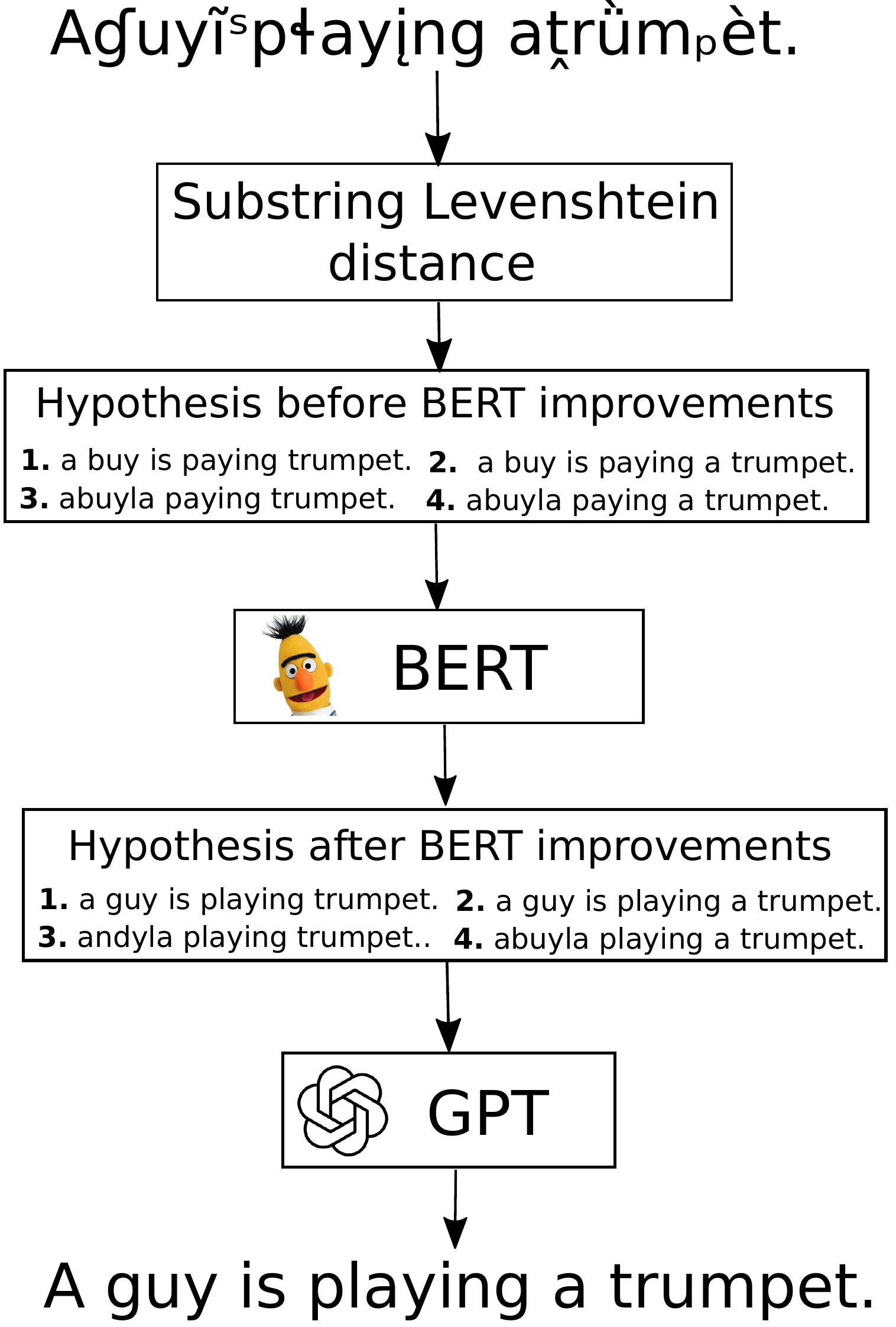}
    \caption{A high-level overview of the processing of an example sentence in our adversarial-defense pipeline. The sentences shown for the hypothesis have been created by choosing the maximum of their associated probability distributions over words.
    }
    \label{fig:overview}
\end{figure}
Often, this is framed as a search problem where the attacker has at least access to model predictions \citep{zang-etal-2020-word}. ``Low-level'' attackers operate on the level of characters and may consist of adversarial typos \citep{Belinkov2017synthetic,Ebrahimi2018OnAdversarial,pruthi-etal-2019-combating,jones-etal-2020-robust} or replacement of characters with similarly looking ones \citep{viper,Li:TextShield}. Such attacks may also be successful when the attacker operates in a blind mode, without having access to model predictions, and they 
are  
arguably more realistic, e.g., in social media. 
However, \citet{pruthi-etal-2019-combating} showed that orthographic attacks can be addressed by placing a spelling correction module in front of a downstream classifier, which may be considered a natural solution to the problem.\footnote{One could argue that such a pipeline solution is not entirely satisfactory from a more theoretical perspective, and that downstream classifiers should be innately robust to attacks in the same way as humans.}

In this work, we apply their approach to the recently proposed  
benchmark \emph{Zéroe} of \citet{benz}, 
 illustrated in Table \ref{tab:adv_attacks}, which provides an array of cognitively motivated orthographic attacks, including missing word segmentation, phonetic and visual attacks. We show that the spelling correction module of \citet{pruthi-etal-2019-combating}, which has been trained on simple typo attacks such as character swaps and character deletions, fails to generalize to this benchmark. This motivates us to propose a novel technique to addressing various forms of orthographic 
 adversaries 
 that does not require to train on the low-level attacks: first, we 
 obtain probability distributions over likely true underlying words from a dictionary using a context-independent extension of the Levenshtein distance; 
 then we use the masked language modeling objective of BERT, which gives likelihoods over word substitutions in context, to refine the obtained probabilities. We iteratively repeat this process to improve the word context from which to predict clean words. 
 Finally, we apply a source text independent language model to produce fluent output text. 

\textbf{Our contributions:}
\textbf{(i)} We empirically show that this approach performs much better than the trained model of \citet{pruthi-etal-2019-combating} on the Zéroe benchmark.  Furthermore, \textbf{(ii)} we also evaluate human robustness on Zéroe and \textbf{(iii)} demonstrate that our iterative approach, which we call \ours{}, sometimes even outperforms human crowd-workers trained via 3-shot learning. 

\begin{table}[!t]
    \centering
    \begin{tabular}{l l}
        \toprule
        \textbf{Attacker} & \textbf{Sentence} \\\midrule
        inner-shuffle & A man is drnviig a car.\\\hline
        full-shuffle & A amn is ginvdir a acr. \\\hline
        disemvowel & A mn is drvng a cr.\\\hline
        intrude & A ma\#n i*s driving a ca\^{}r. \\\hline
        keyboard-typo & A mwn is dricing a caf. \\\hline
        natural-typo & A wan his driving as car. \\\hline
        truncate & A man is drivin a car. \\\hline
        segmentation & Aman isdriving a car. \\\hline
        phonetic & Ae man izz dreyvinn a cahar.\\\hline
        visual & \includegraphics[width=0.45\columnwidth,valign=b]{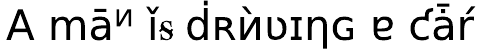}
        \\\bottomrule
    \end{tabular}
    \caption{Examples for the adversarial attacks from the Zéroe benchmark. The phonetic and visual examples show our modified implementations (see appendix \ref{app:vis_phon}).}
    \label{tab:adv_attacks}
\end{table}

%% file: related.tex
\section{Related work}

\citet{zeng2020openattack} classify adversarial attack scenarios in terms of the \emph{accessibility} of 
the victim model to the attacker:\footnote{Another recent survey of adversarial attacks in NLP is provided by \citet{roth2021token}.} \emph{white-box} attackers \citep{Ebrahimi2018Hotflip} have full access to the victim model including its gradient to construct adversarial examples. In contrast, \emph{black-box} attackers have only limited  knowledge 
of the victim models: score- \citep{Alzantot2018Generating,Jin:2020} and decision-based attackers \citep{Ribeiro2018Semantically} require access to the victim models' prediction scores (classification probabilities) and final decisions (predicted class), respectively. 
A score-based black-box attacker of particular interest in our context is BERT-ATTACK \citep{bert-attack}. BERT-ATTACK uses the masked language model (MLM) of BERT to replace words with other words that fit the context. BERT-ATTACK is related to our approach because it uses BERT's MLM in an attack-mode while we use it in defense-mode. Further, in our terminology, BERT-ATTACK is a high-level attacker, while we combine BERT with an edit distance based approach to restore low-level adversarial attacks.  
\emph{Blind} attackers make fewest assumptions and have no knowledge of the victim models at all. Arguably, 
they are most realistic, e.g., in the context of online discussion forums and other forms of social media where users may not know which model is employed (e.g.) to censor toxic comments and users may also not  
have (large-scale) direct access to model predictions. 

In terms of blind attackers, \citet{viper} design the visual perturber VIPER which replaces characters in the input stream with visual nearest neighbors, an operation to which humans are seemingly very robust.\footnote{Basing text processing on visual properties was also recently explored in \citet{wang2020word} and \citet{salesky:2021}.} \citet{benz} propose a canon of 10 cognitively inspired orthographic character-level blind  attackers. We use this benchmark, which is illustrated in Table \ref{tab:adv_attacks}, 
in our application scenario. 
While \citet{viper} and \citet{benz} are only moderately successful in defending against their orthographic attacks with \emph{adversarial learning} \citep{Goodfellow2014Explaining} (i.e., including perturbed instances at train time), 
\citet{pruthi-etal-2019-combating} show that placing a word recognition (correction) module in front of a downstream classifier may be much more effective. They use a correction model trained to recognize words corrupted by random adds, drops, swaps, and keyboard mistakes. 
\citet{zhou-etal-2019-learning} also train on the adversarial attacks (insertion, deletion, swap as well as word-level) against which they defend. 
In contrast, we show that an \emph{untrained} attack-agnostic iterative model based on BERT may perform competitively even with humans (crowd-workers) and that this correction module may further be improved by leveraging attack-specific knowledge. \citet{jones-etal-2020-robust} place an encoding module---which should map orthographically similar words to the same (discrete) `encoding'---before the downstream classifier to improve robustness against adversarial typos. However, in contrast to \citet{pruthi-etal-2019-combating} and \ours{}, their model does not 
restore the attacked sentence to its original form so that it is less desirable in situations where knowing the underlying surface form may be relevant (e.g., for human introspection or in tasks such as spelling normalization). 

In contemporaneous work, \citet{hu2021misspelling} 
use BERT for masked language modeling together with an edit distance to correct a misspelled word in a sentence. They assume a single  
misspelled word that they correct by selecting from a set of edit distance based hypotheses using BERT. In contrast, in our approach we assume that multiple or even all words in the sentence have been attacked using adversarial attacks and that we do not know which ones. Then, we use an edit distance and integrate its results probabilistically with context information obtained by BERT,  
rather than using edit distance only for candidate selection.  

%% file: method.tex
\section{Methods}
Our complete model, which is outlined in Figure \ref{fig:overview} on a high level, has three intuitive components. The first component is context-independent and tries to detect the tokens in a sentence from their given (potentially perturbed) surface forms. This makes sense, since we assume orthographic low-level attacks on our data. The second component uses context, via masked language modeling in BERT, to refine the probability distributions obtained from the first step. The third component uses a language model (in our case, GPT) to make a choice between multiple hypotheses. 
In the following, we describe each of the three components. 

\subsection{Context-independent probability}
\label{sec:lev}
In the first step of our sentence restoration pipeline, we use a modified Levenshtein distance to convert the sentence into a list of probability distributions over word-piece tokens from a dictionary $D$. For the dictionary, we choose BERT's \cite{bert} default word-piece dictionary. 

We begin by splitting the attacked sentence $S$ at spaces into word tokens $\tilde{w}_i$. However, to 
be able to use 
our word-piece dictionary $D$, we need to find the appropriate segmentation of the tokens into word-pieces. 

\paragraph{Modified Levenshtein distance.} We developed a modified version of the Wagner–Fischer algorithm \cite{wagner_fischer}
that calculates a Levenshtein distance to substrings of the 
input string 
and keeps track of start as well as end indices of matching substrings. 
For each $\tilde{w}_{i}$ in $S$,  
this algorithm (which is described in Appendix \ref{wagner-fischer}) calculates 
the substring Levenshtein distance \emph{dist} to every word-piece $w_{d}$ in $D$.
\paragraph{Segmentation hypothesis.} We store the computed distances $\textit{dist}(\tilde{w}_{i},w_d)$ in a  dictionary $C_i$ that maps each start-index $s$ and end-index $e$ to a list of distances, i.e., $C_i$ associates 
\begin{align*}
    (s,e)\mapsto \bigl(\textit{dist}(\tilde{w}_{i},w_{d})\bigr)_{w_{d}\in D'}
\end{align*}
Here, $D'$ 
selects the subset of all word-pieces in $D$ that match $\tilde{w}_{i}$ at the substring between $s$ and $e$. 
Using  $C_i$, we can then perform a depth-first search to 
compose $\tilde{w}_{i}$ from start and end-indices in $C_i$. For example, a 10 character word $\tilde{w}_{i}$ could be segmented into two words-pieces that match the substrings from positions 1-5 and 6-10,  respectively, or a single word that matches from 1-10. Let $\mathbf{c}_{i}$ be the 
set of all segmentations of  
$\tilde{w}_{i}$ from start and end indices. 
For example, $\mathbf{c}_i$ could be $\bigl\{\bigl((1,5),(6,10)\bigr),\,\bigl((1,10)\bigr)\bigr\}$. 
For each segmentation $\mathbf{c}_{i,\alpha}\in \mathbf{c}_i$, 
we then 
calculate 
a \emph{total distance} $d(\mathbf{c}_{i,\alpha})$ as a sum of the minimum distances of all parts:
\begin{align}\label{eq:tdist}
    d(\mathbf{c}_{i,\alpha}) = \sum_{k=1}^{\text{len}(\mathbf{c}_{i,\alpha})}\min(C_i[\mathbf{c}_{i,\alpha,k}])
\end{align}

\begin{figure}[!t]
    \centering
    \includegraphics[width=1\columnwidth]{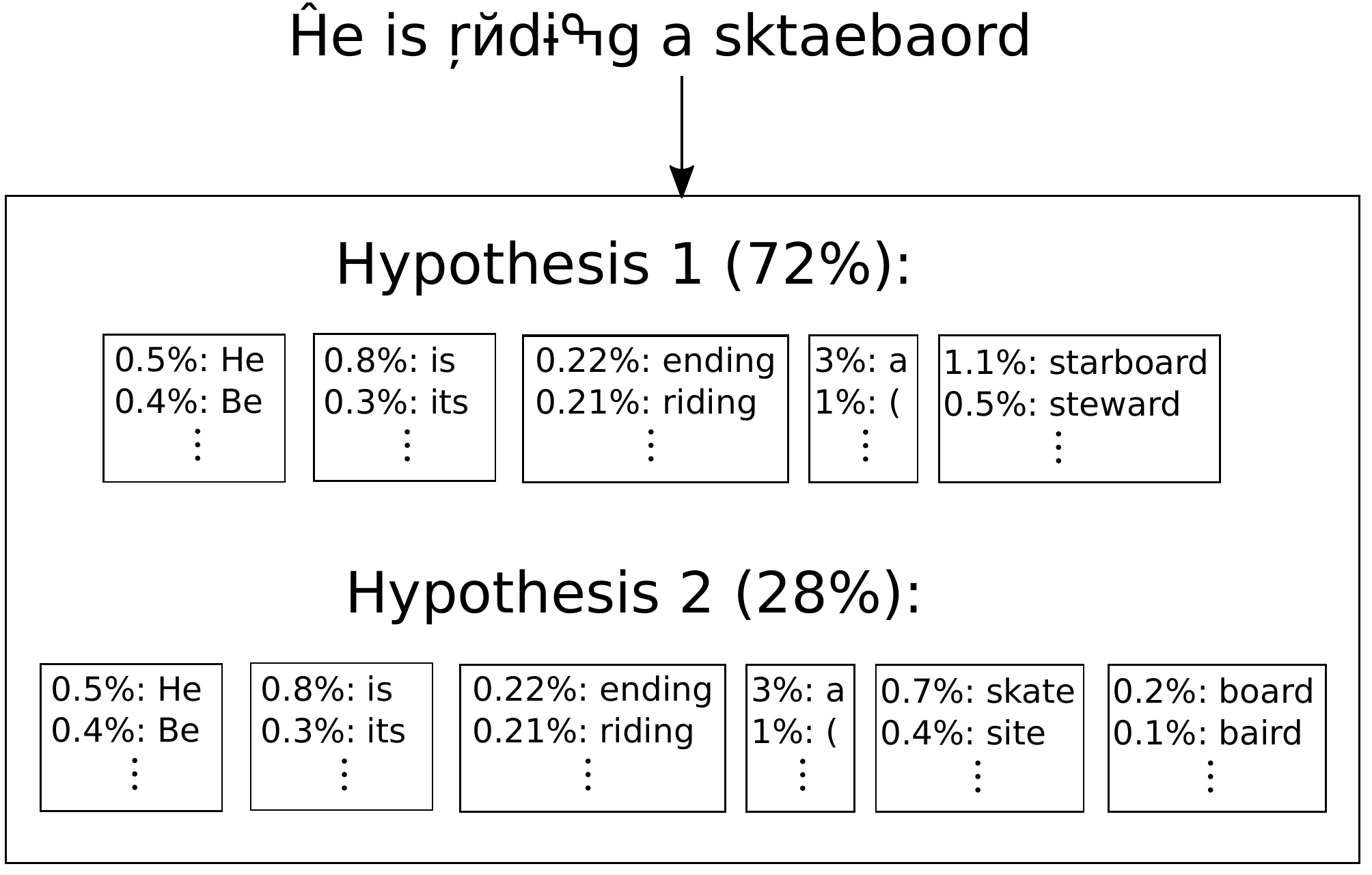}
    \caption{A context independent probability distribution over words calculated for an example input sentence. There are multiple segmentation-hypothesis associated with the sentence that each consist of a sequence of probability distributions over word-tokens.}
    \label{fig:prob_no_context}
\end{figure}

Using the total distances to segment each token $\tilde{w}_{i}$, we can now create hypotheses $\mathbf{H}$ about how the whole sentence $S$ consisting of $n$ tokens should be segmented into word-pieces. For this, we calculate the Cartesian product between the sets of possible segmentations for each word $\tilde{w}_{i}$, $i=1,\ldots,n$:
\begin{align*}
\mathbf{H} = \mathbf{c}_1\times \mathbf{c}_2\times \cdots \times \mathbf{c}_n
\end{align*}
We set 
the \emph{loss} 
of one hypothesis $\mathbf{h}=(\mathbf{c}_{1,\alpha_1},\ldots,\mathbf{c}_{n,\alpha_n}) \in \mathbf{H}$  
as the sum of the total distances of its parts that we calculated in Eq.~\eqref{eq:tdist}
\begin{align*}
\textit{loss}(\mathbf{h}) 
= \sum_{i=1}^n 
d(\mathbf{c}_{i,\alpha_i})
\end{align*}
By evaluating the softmax on the negative total distances of the hypothesis, we calculate probabilities if a hypothesis $\mathbf{h}_v\in \mathbf{H}$ is equal to the true (unknown) segmentation $\mathbf{h}^{*}$ of the $n$ tokens: 
\begin{equation}
\label{eq:seg_prob}
\begin{split}
    \mathbf{d}_\mathbf{H} &= (\textit{loss}(\mathbf{h}))_{\mathbf{h}\in \mathbf{H}}\\
    \mathbf{P}(\mathbf{h}_v{=}\mathbf{h}^{*}\mid S) &= [\text{softmax}(-\mathbf{d_H})]_v
\end{split}
\end{equation}
We will refer back to these probabilities in \S\ref{sec:gpt}.
\paragraph{Word probability distributions.}
In a hypothesis $\mathbf{h}\in\mathbf{H}$, a token  
$\tilde{w}_{i}$ has a single segmentation of start and end indices associated with it, $\mathbf{c}_{i,\alpha}$. For all start- and end-indices $(s,e)$, $C_i[s,e]$ stores the distances of the words that match $\tilde{w}_i$ between $s$ and $e$. Let $D'$ again be the dictionary containing all those words. Let $w_{d}$ 
be a word-piece in $D'$ 
and let $w^{*}\in D'$ be the true match for the substring between $s$ and $e$ of $\tilde{w}_i$. 
Then, we can compute a context-independent probability that $w_{d}$ is equal to $w^{*}$, by evaluating the softmax on the negative distances stored in $C_i$:
\begin{align*}
    \mathbf{P_h}(w_{d}=w^{*}\mid \tilde{w}) = [\text{softmax}(-C_i[s,e])]_d
\end{align*}
When we do this for all words in $\mathbf{h}$ and concatenate the results, we get a vector $\mathbf{V_h}$ of probability distributions over dictionary word-pieces. This is illustrated 
in Figure \ref{fig:prob_no_context}. We introduce the following notation to select a probability distribution based on its index in $\mathbf{h}$ using the subscript $j$:
\begin{align*}
    \mathbf{P}_{\mathbf{h},j}(w_d=w^{*}\mid \tilde{w}) \coloneqq \mathbf{V}_{\mathbf{h},j}
\end{align*}

\paragraph{Domain-specific distance.}In the remainder, we will refer to the way of calculating the substring distance
as described above as \emph{\basedist{}}. 
Beyond this, we also aim to leverage domain-specific knowledge. 
We refer to such an augmented distance as the
\emph{\fulldist{}} distance $\textit{dist}_M$. 
Here, we modify the operation costs in the substring Levenshtein distance in certain situations.
\begin{enumerate}
    \item Edit distance is reduced for visually similar characters. This builds on visual character representations \citep{viper}.
    See appendix \ref{vis-sim} for details.
    \item Addressing intruder attacks, we reduce deletion costs depending on the frequency $f$ of the character in the source word. 
    Our assumption is 
    that the same intruder symbol may be repeated in one word.
    Thus, we decay the cost exponentially for increasing frequency using the formula $0.75^{f-1}$. 
    \item Vowel insertion cost is reduced to 0.3 for words that contain no vowels.
\end{enumerate}
To address letter-shuffling, we additionally compute an anagram distance of how close the attacked word $\tilde{w_i}$ is to being an anagram to the dictionary word $w_d$. Let $m$ be the number of characters that are in one of the two words, but not in the other. Then, our anagram distance $\textit{dist}_{\text{A}}$ computes to  
\begin{align*}
    \textit{dist}_{\text{A}}(\tilde{w_i},w_d) = 2m+1
\end{align*}
When two words are permutations of each other, the anagram distance is minimal and otherwise it increases linearly in the number of different characters between the two words. 
We then take the minimum of the anagram distance and the substring Levenshtein distance with modified operation costs $\textit{dist}_\text{M}$ to obtain the 
\emph{\fulldist{}} $\textit{dist}_\text{F}$:
\begin{align*}
    \textit{dist}_{\text{F}}(\tilde{w_i},w_d) = \min(\textit{dist}_{\text{A}}(\tilde{w_i},w_d),\textit{dist}_{\text{M}}(\tilde{w_i},w_d))
\end{align*}
\subsection{Context-dependent probability using BERT}
\label{sec:bert}
\begin{figure}[!t]
    \centering
    \includegraphics[width=1\columnwidth]{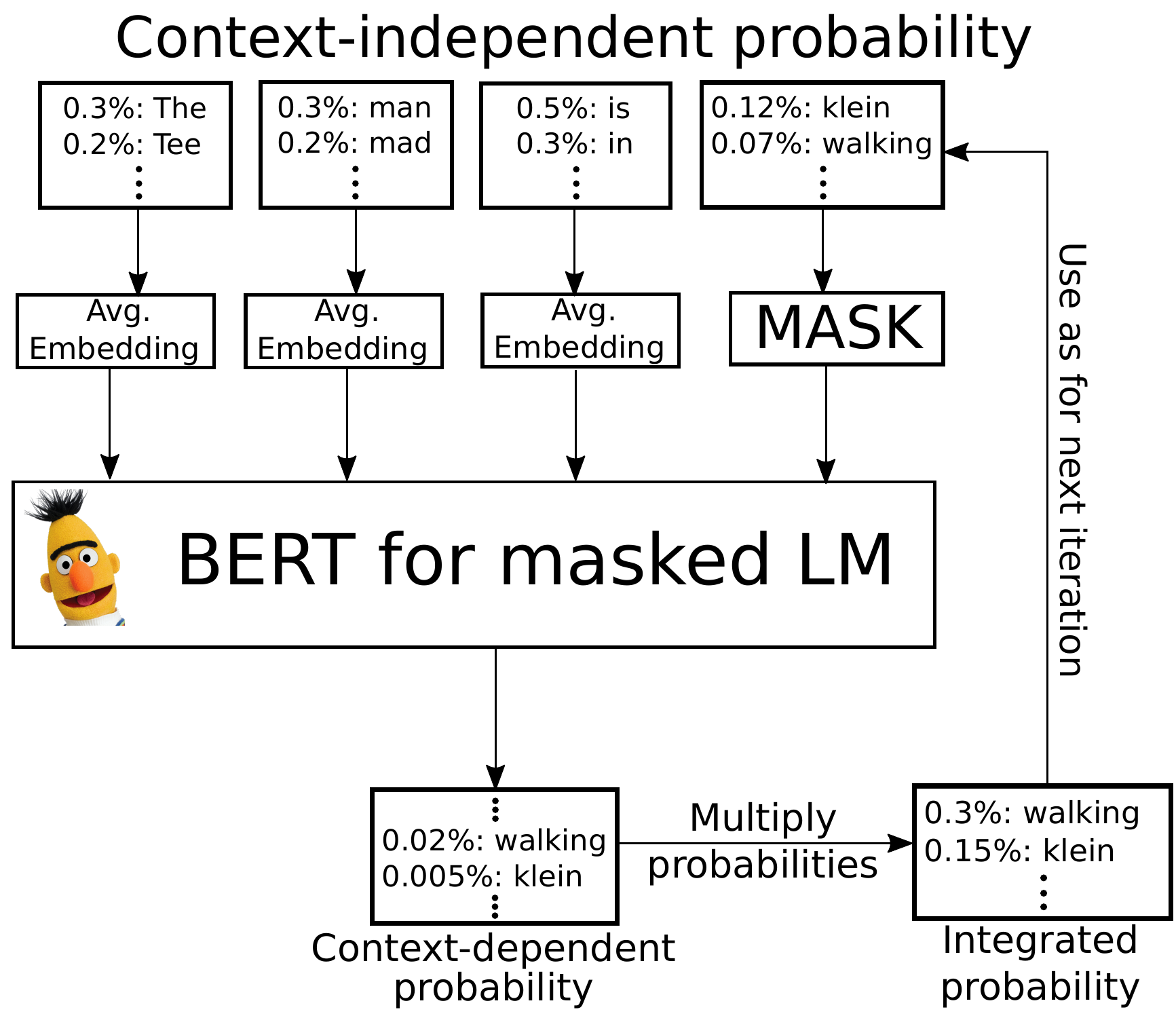}
    \caption{Iterative, context-based improvements of the word predictions using BERT for masked LM. Each iteration, a different token will be masked. We calculate context-dependent probabilities using Eq.  \eqref{eq:softmax} and integrate them with our context-independent probabilities in Eq. \eqref{eq:update}.}
    \label{fig:bert_posterior}
\end{figure}
In the following, we describe the context-based improvement for a single hypothesis $\mathbf{h}\in \mathbf{H}$. In Figure \ref{fig:bert_posterior}, the whole process is illustrated for an example sentence. The number of required iterations should scale linearly with the amount of tokens in the hypothesis, so we perform $2\cdot|\mathbf{h}|$ iterations in total. To perform one improvement iteration, we perform the following steps:

\noindent 1) Select an index $j$, of a token, that will be masked for this iteration. 

\noindent 2) For the next part, we slightly modify BERT for masked LM. 
Instead of using single tokens as inputs as in BERT, we want to use our context-independent probability distributions over word-piece tokens. 
Thus, for each token $w_{\mathbf{h},j}$ in $\mathbf{h}$, we embed all relevant tokens $w_d$ from the context-independent process described above 
using BERT's embedding layer and combine them into a weighted average embedding using weights $\mathbf{P}_{\mathbf{h},j}(w_{d}=w^{*}\mid \tilde{w})$. 

\noindent 3) We now bypass BERT's embedding layer and feed the weighted average embeddings and the embedding for the mask token directly into the next layers of BERT\footnote[1]{Although BERT has only been trained on the single token embeddings, we empirically found that feeding in averaged embeddings produces very sensible results.}. As a result, BERT provides us with a vector of scores $\mathbf{S_{\text{BERT}}}$ 
for how well the words from the word-piece dictionary $D$ fit into the position of the masked word. 

\noindent 4) By applying the softmax on these scores, we obtain a new probability distribution over word-pieces which is dependent on the context $c$ of the token at position $j$:
\begin{equation}\label{eq:softmax}
    \mathbf{P}_{\mathbf{h},j}(w_{d}=w^{*}\mid c)=[\text{softmax}(\mathbf{S_{\text{BERT}}})]_d
\end{equation}
5) 
We make the simplifying assumption that each word is attacked independently from the other words. 
Thus,  
the context $c$ is independent of the attack on the word $\tilde{w}$. This means that the following equality holds:
\begin{equation}\label{eq:update}
\begin{split}
    \mathbf{P}_{\mathbf{h},j}&(w_{d}=w^{*}\mid \tilde{w},c)  =\\ &\mathbf{P}_{\mathbf{h},j}(w_{d}=w^{*}\mid \tilde{w})\mathbf{P}_{\mathbf{h},j}(w_{d}=w^{*}\mid c)
    \end{split}
\end{equation}
6) We go back to step 1) and use $\mathbf{P}_{\mathbf{h},j}(w_d=w^{*}\mid \tilde{w},c)$ from Eq.~\eqref{eq:update} instead of $\mathbf{P}_{\mathbf{h},j}(w_d=w^{*}\mid \tilde{w})$ to create the average embedding at position $j$. 
\subsection{Selecting the best hypothesis with GPT}
\label{sec:gpt}
\begin{figure}[!t]
    \centering
    \includegraphics[width=1\columnwidth]{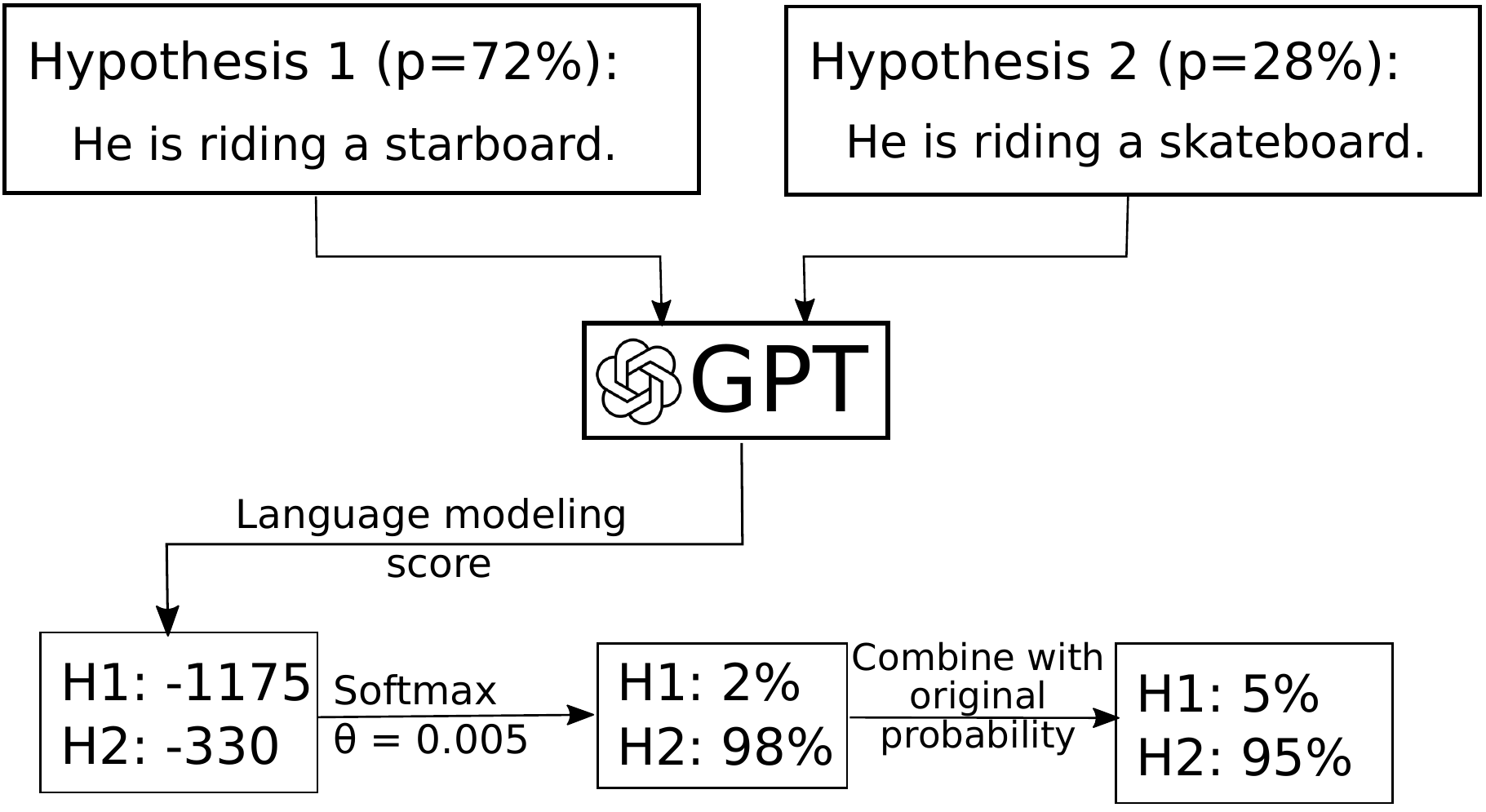}
    \caption{An example of how we use OpenAI GPT to decide on which hypothesis to choose as our final sentence prediction. The original probability of the segmentation hypothesis calculated in Eq. \eqref{eq:seg_prob} is multiplied with a probability calculated from the language modeling score using Eq. \eqref{eq:mult_gtp_probs}.}
    \label{fig:gpt}
\end{figure}

After performing the context-based improvements, we are left with multiple hypothesis $\mathbf{h}\in \mathbf{H}$. Each of them has a hypothesis probability $\mathbf{P}(\mathbf{h}{=}\mathbf{h^*}\mid S)$ and a 
list of word-piece probabilities of length $|\mathbf{h}|$ over dictionary words associated with it. Now, we finally collapse the probability distributions by taking the argmax to form actual sentences $S_\mathbf{h}$: 
\begin{equation}
\begin{split}
    w_{\mathbf{h},j} &\gets \argmax_{w_d}(\mathbf{P}_{\mathbf{h},j}(w_d=w^{*}\mid \tilde{w},c))\\
    S_\mathbf{h} &\gets (w_{\mathbf{h},j})_{j=1}^{|\mathbf{h}|}
    \end{split}
\end{equation}
This allows us to use GPT \cite{gpt} to calculate a language modeling 
(perplexity)
score  $\text{LM}_{S_\mathbf{h}}$ for each sentence. Using softmax, we  
again transform these scores into a probability distribution that describes the probability of a segmentation hypothesis $\mathbf{h}_v\in \mathbf{H}$ being the correct segmentation $\mathbf{h^*}$, based on the restored sentences $S_\mathbf{H}$:
\begin{align*}
    S_\mathbf{H} &\gets (S_\mathbf{h})_{\mathbf{h}\in\mathbf{H}}\\
    \text{LM}_{S_\mathbf{H}} &\gets (\text{LM}_{S_\mathbf{h}})_{\mathbf{h}\in\mathbf{H}}\\
    \mathbf{P}(\mathbf{h}_v{=}\mathbf{h^*}\mid S_\mathbf{H}) &= [\text{softmax}(\text{LM}_{S_\mathbf{H}})]_v
\end{align*}
The original probability $\mathbf{P}(\mathbf{h}_v{=}\mathbf{h^*}\mid S)$ assigned to each hypothesis is only based on the results of the Levenshtein distance 
for the attacked sentence $S$. Thus, as 
$\mathbf{P}(\mathbf{h}_v{=}\mathbf{h^*}\mid S)$ 
only depends on the character-level properties of the attacked sentence 
and $\mathbf{P}(\mathbf{h}_v{=}\mathbf{h^*}\mid S_\mathbf{H})$ only depends on the semantic properties of the underlying sentences,  
it makes sense to assume that these distributions are independent. This allows us to simply multiply them, to get a probability distribution that captures semantic as well as character-level properties:
\begin{equation}
\label{eq:mult_gtp_probs}
\begin{split}
    \mathbf{P}(\mathbf{h}_v{=}\mathbf{h^*}\mid S_\mathbf{H},S) &=\\ \mathbf{P}(\mathbf{h}_v{=}&\mathbf{h^*}\mid S_\mathbf{H})\mathbf{P}(\mathbf{h}_v{=}\mathbf{h^*}\mid S)
\end{split}
\end{equation}
In Figure \ref{fig:gpt}, we visualize the above described process for a specific example with only 2 hypotheses.

%% file: experiments.tex
\section{Experimental Setup}\label{sec:experiments}
To obtain adversarially attacked sentences against which to defend, we use the \citet{benz} benchmark Zéroe of low-level adversarial attacks. This benchmark contains implementations for a wide range of
cognitively inspired 
adversarial attacks such as letter shuffling, disemvoweling, phonetic and visual attacks.
The attacks are parameterized by a perturbation probability $p\in[0,1]$ that controls how strongly the sentence is attacked.

We decided to slightly modify two of the attacks in Zéroe, the phonetic and the visual attacks. 
On close inspection, we found the phonetic attacks to be too weak overall, with too few perturbations per word. The visual attacks in Zeroé are based on pixel similarity which is similar to the visual similarity based defense in our \emph{domain-specific} model. Thus, to avoid attacking with the same method we defend with, we decided to switch to a description based visual attack model (DCES), just like in the original paper \citep{viper}.\footnote{Using  description based defense and pixel based attacks would have been possible just as well, but we believe doing it reversely is consistent with the original  specification in \citet{viper}.}
Our modifications are described in Appendix \ref{app:vis_phon}.

\paragraph{Evaluation}
Instead of evaluating on a downstream task, we evaluate on the task of restoring the original sentences from the perturbed sentences. This 
allows us to easier compare to human performances. 
It also provides a more difficult test case, as a downstream classifier may infer the correct solution even with part of the input destroyed or omitted. Finally, being able to correct the input is also important when the developed tools would be used for humans, e.g., in spelling correction. 

We evaluate the similarity of the sentences to the ground-truth sentences with the following metrics:
\begin{figure*}[!t]
    \centering
    \includegraphics[width=0.95\textwidth]{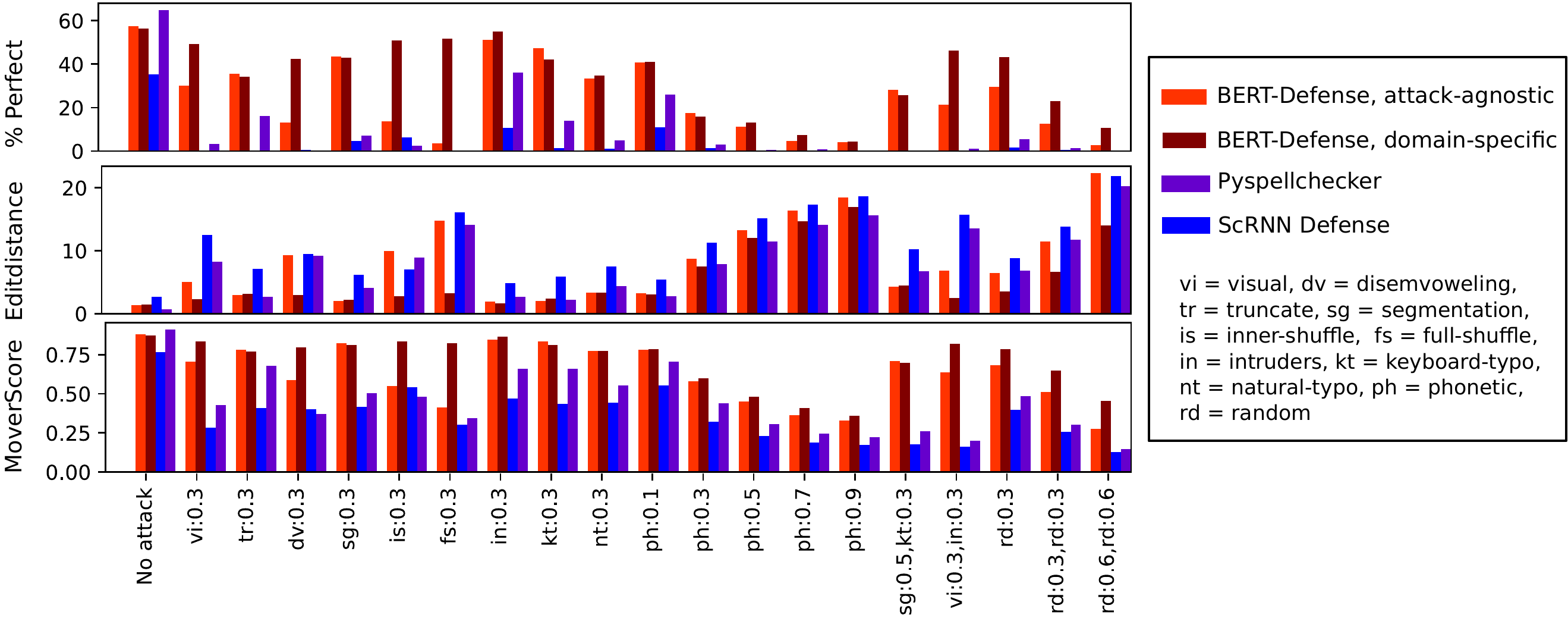}
    \caption{Comparison between \ours{}, and the two baseline adversarial defense tools ``pyspellchecker'' and ``ScRNN defense''. The x-labels describe the attack and perturbation level the sentences were attacked with, before applying on of the adversarial defense methods. For conditions with two attack types, the perturbations were applied in order. For edit distance, lower is better. For the other metrics, higher is better. Exact values for the results are included in the appendix in Table \ref{tab:exact_data}.}
    \label{fig:evaluations_baseline}
\end{figure*}
\begin{enumerate}[topsep=5pt,itemsep=0pt,leftmargin=*]
    \item \textbf{Percent perfectly restored} (PPR). The percent of sentences that have been restored perfectly. 
    This is a coarse-grained sentence-level measure. 
    \item \textbf{Editdistance}. The Levenshtein (edit) distance measures the number of insertions, deletions, and substitutions necessary (on character-level) to transform one sequence into another. 
    \item \textbf{MoverScore} \cite{MoverScore}. MoverScore measures the semantic similarity between two sentences using BERT. It has been show to correlate highly with humans as a semantic evaluation metric.
\end{enumerate}
For all of the metrics, letter case was ignored.

\paragraph{Attack scenarios.}
We sampled 400 sentences from the GLUE \cite{glue} STS-B development dataset for our experiments. 
We use various attack scenarios to attack the sentences:
\begin{itemize}[topsep=5pt,itemsep=0pt,leftmargin=*]
\item[i)] Each of the attack types of the Zéroe benchmark (see Table \ref{tab:adv_attacks}). 
We set $p=0.3$ throughout. 
\item[ii)] To evaluate how higher perturbation levels influence restoration difficulty, we create 5 attack scenarios for one attack scenario (we randomly chose phonetic attacks) with perturbation levels $p$ from $0.1$ to $0.9$.
\item[iii)] We add  
combinations of attacks: these are performed by first attacking the  sentence with one attack and then with another. 
\item[iv)]  In Random attack scenarios (rd:0.3, [rd:0.3,rd:0.3], [rd:0.6,rd:0.6]), one or two attack types from the benchmark are randomly chosen for each sentence. 
These constitute stronger attack situations and may be seen as more challenging test cases. 
\end{itemize}
Each of the 19 attack scenarios is applied to all 400 sentences 
individually to create 19 test cases of attacked sentences.
\paragraph{Baselines and upper bounds.}
To evaluate how well \ours{} restores sentences, we compare its sentence restoration ability to two baselines: 
(a) the \emph{Pyspellchecker} \cite{pyspellchecker}, a simple spellchecking algorithm that uses the Levenshtein distance and word frequency to correct errors in text; 
(b) ``ScRNN defense'' from the \citet{pruthi-etal-2019-combating} paper. This method uses an RNN that has been trained to recognize and fix character additions, swaps, deletions and keyboard typos. 
Further, as we use Zeroé, a cognitively inspired attack benchmark 
supposed to fool machines but not humans, it is especially interesting to see how \ours{} compares to human performance. Thus, (c) we include human performance, 
obtained from a crowd-sourcing experiment on Amazon Mechanical Turk (AMT). Note that humans are often considered upper bounds in such settings. 
\paragraph{Human experiment.}
Twenty-seven subjects were recruited using AMT (21 male, mean age 38.37, std age 10.87) using PsiTurk \cite{psiTurk}. Participants were paid \$3 plus up to \$1 score based bonus (mean bonus 0.56, std bonus 0.40) for restoring about 60 adversarially attacked sentences. 
The task took on average 43.9 minutes with a standard deviation of 20.1. 
Twenty of the subjects where native English speakers, 
seven where non-native speakers. 
The two groups did not  
significantly differ regarding their edit distances to the true underlying sentences (unequal variance t-test, $p=.85$).

We sampled 40 random sentences from nine of our 
attack scenarios plus 40 random (non-attacked) sentences from the original document. 
Each sentence was restored by four different humans. 
The whole set of 1600 sentences (10 scenarios times 40 sentences each times 4 repetitions) was then randomly split into 27 sets of about 60 sentences. 
No split contained the same sentence multiple times. Each of the 27 participants got one of these sets assigned. 
After a short instruction text, the participants where shown three examples of how to correctly restore a sentence (``3-shot learning''). Then they were shown the sentences in their set sequentially and entered their attempts at restoring the sentences into a text-field. 

%% file: results.tex
\section{Results and Discussion}
\begin{figure*}[!h]
    \centering
    \includegraphics[width=0.8\textwidth]{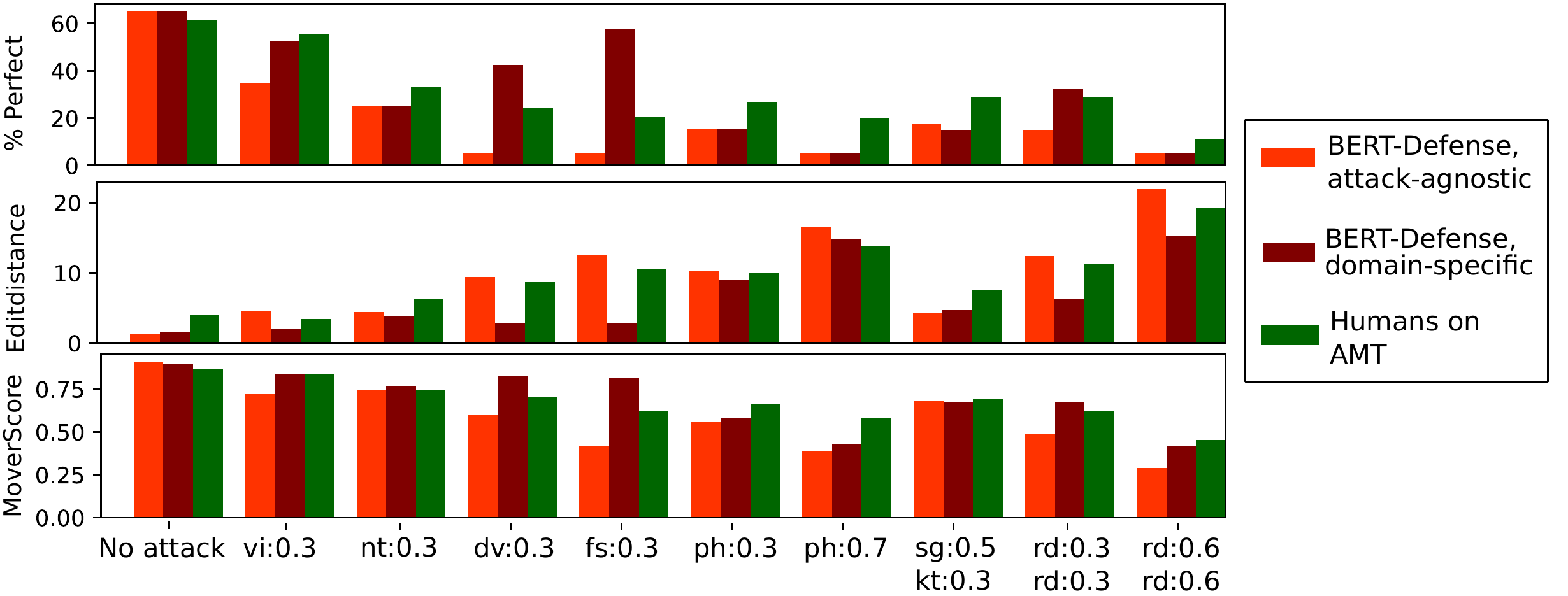}
    \caption{Comparison between \ours{} and humans on Amazon Mechanical Turk.}
    \label{fig:evaluations_human}
\end{figure*}

\paragraph{Comparison with baselines.}
Figure \ref{fig:evaluations_baseline} visualizes the results (full results are in the appendix).
$\text{BD}_{\text{agn}}$ (\ours{}, \basedist{}) significantly outperforms both baselines regarding \emph{MoverScore} and \emph{PPR} for all random attack scenarios (p $\ll 0.01$, equal variance t-test). However, only $\text{BD}_{\text{spec}}$ (\ours{}, \fulldist{}) achieves a lower edit distance than the baselines. This discrepancy between the measures is explained by the fact that, by taking context into account, \ours{} searches for the best restoration in the space of sensible sentences, while \emph{Pyspellchecker} searches the best restoration for each word individually. Although \emph{ScRNN defense} uses an RNN and is able to take context into account, we found that it also mainly seems to restore the words individually and  rarely produces grammatically correct sentences for strongly attacked inputs.
Table \ref{tab:failure_cases}, which illustrates failure cases of all models, supports this. 
In the failure case when \ours{} fails to recognize the correct underlying sentence, 
\ours{} outputs a mostly different sentence that usually does make some sense, but has little 
in common with the ground-truth sentence. This results in much higher edit distances than the failure cases of the baselines  
which produce grammatically wrong sentences, while restoring individual words the best they can (this sometimes means not trying at all). Interestingly, humans tend to produce similar failure cases as \ours{}.

When comparing the performance 
on specific attacks, 
we see 
a consistent margin of about 0.2 MoverScore and 15-35 percentage points PPR between $\text{BD}_{\text{agn}}$ 
and the baselines
across all attacks. 
Exceptions include inner-shuffle, for which ScRNN-Defense is on par with $\text{BD}_{\text{agn}}$ and segmentation attacks, which hurt the performance of the baselines far more than the performance of \ours{}, which includes segmentation hypothesis as an essential part of its restoration pipeline. For $\text{BD}_{\text{spec}}$, we see gains 
for attacks 
where we leverage 
domain-specific knowledge. The biggest gains of around 0.25 MoverScore are achieved against full-shuffle, inner-shuffle and disemvoweling attacks.

\begin{table*}[!htb]
    \centering
    {\small
    \begin{tabular}{l|lll}
         \toprule
         \textbf{Attacked Sentence} & \textbf{ScRNN} & \textbf{$\text{BD}_{\text{agn}}$}\\
         \midrule
         To lorge doog's wronsing in sum grass. & to lorge doog's wronsing in sum grass. & two large dogs rolling in the grass.	\\
         Two large dogs runningin some grass. &two large dogs runningin some grass.&two large dogs running in some grass. \\
         Tw large dogs rnnng in some grss.&throw large dogs running in some grss.&two large dogs running in some grass.\\
         Two larg dog runnin in some grass.&two larg dog runnin in some grass.&a large dog running in some grass.\\
         Twolarge dogs running income graas.&twolarge dogs running income graas.&two large dogs running into grass.\\
         To lrg doog's rntng in sm gras ..&to long dogs ring in sm gras.&to the dogs running in the grass.\\
         \bottomrule
    \end{tabular}
    }
    \caption{Various illustrative attacks on the sentence ``Two large dogs running in some grass.'' and restorations by ScRNN and BD$_{\text{agn}}$. The attacked sentences are attacked with the following attacks (top to bottom): Phonetic-0.7, Segmentation-0.3, Disemvowel-0.3, Truncate-0.3, Segmentation-0.5 \& Keyboard-Typo-0.3, Random-0.3 \& Random-0.3 (the last two are double attacks).}
    \label{tab:morecases}
\end{table*}

In the \emph{No attack} condition, we checked if the adversarial defense methods introduce mistakes when presented with clean sentences. 
Indeed, all models introduce some errors:  all three evaluation metrics show that \ours{} introduces a few more errors than \emph{Pyspellchecker} but less than \emph{ScRNN defense}. 

\begin{table}[!t]
    \centering
    \noindent
\begin{tabularx}{\linewidth}{@{}>{\bfseries}p{2.8cm}@{\hspace{.5em}}X@{}}
\toprule
    Ground-truth & china gives us regulators access to audit records \\
    Attacked (rd:0.6,rd:0.6) &  hainc gcive us regulafors essacf to tufai rsxrdeo \\ \midrule
    Bert-Defense (\basedist{}) & haine gave us regulators space to turn us over \\
    Human & hain gives us regulators escape to dubai suborder \\
    ScRNN Defense & hainc give us regulafors essacf to tufai rsxrdeo \\
    Pyspellchecker & hain give us regulators essay to tufa rsxrdeo\\
    \bottomrule
\end{tabularx}
    \caption{Failure cases for \ours{}, humans and the baseline methods. Note that in the failure case, \ours{} and Humans restore sentences that are grammatically correct, but are mostly different from the ground-truth. On the other hand, Pyspellchecker and ScRNN Defense \citep{pruthi-etal-2019-combating} either refuse to try at all for strongly attacked words or create grammatically nonsensical sentences. 
    }
    \label{tab:failure_cases}
\end{table}
\paragraph{Comparison with humans.}
As stated before, 
we evaluate human performance  
on 40 random sentences for each of nine attacks and the no attack condition (see appendix). 
For each of the sentences, we obtain  restorations from 4 crowd-workers. For each attack scenario, we evaluate our metrics on all restorations of these 40 sentences and averaged the results. 
The results on the 40 attacked sentences are shown in Figure \ref{fig:evaluations_human}. 
While $\text{BD}_{\text{agn}}$ performs slightly worse than humans, $\text{BD}_{\text{spec}}$ 
matches human performance with respect to all three evaluation metrics. Regarding performance on specific attacks, humans are still better than \ours{} when it comes to defending phonetic attacks, while they have a hard time defending full-shuffle attacks. The evaluations for the \emph{No attack} setting reveal that the crowd-workers in our experiment do make quite a few copying mistakes. In fact, they introduce slightly more mistakes than \ours{}.
\paragraph{Ablation Study.}
\begin{figure}[!h]
    \centering
    \includegraphics[width=0.45\textwidth]{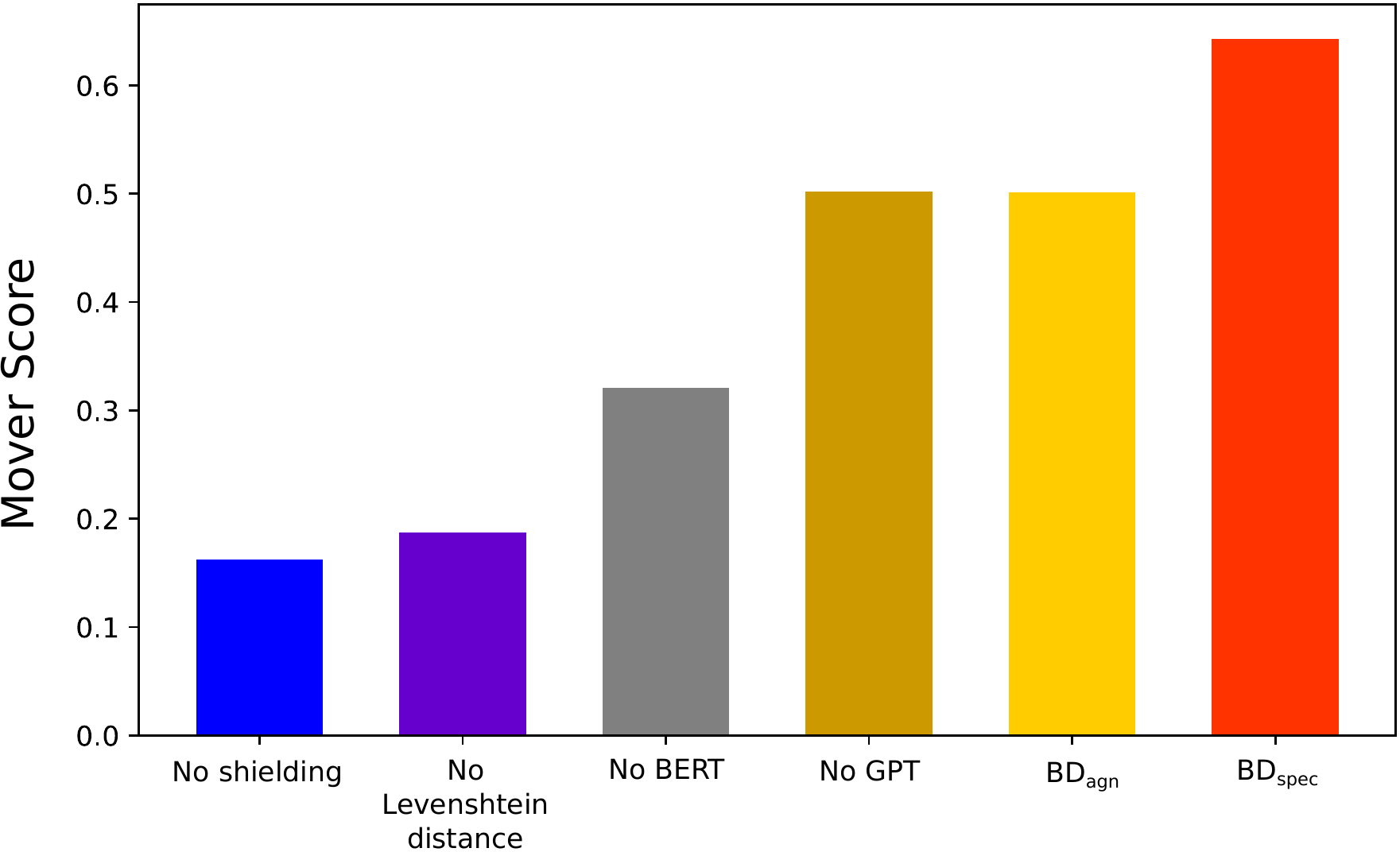}
    \caption{Ablation study for BERT-Defense. The MoverScore metric is shown for BERT-Defense with  exactly one single component left out, respectively, on the rd:0.3,rd:0.3 attack. For comparison, we also show the MoverScore without shielding and after shielding with $\text{BD}_{agn}$ or $\text{BD}_{spec}$ using all components.}
    \label{fig:ablation}
\end{figure}
We perform an ablation study to asses the contribution of each individual component of BERT-Defense. For the \emph{No Levenshtein distance} condition, we created the context-independent probability distribution by setting the probability of known words (words in the dictionary) in the attacked dataset to one and using a uniform random distribution for all unknown words. When using BERT-Defense without BERT, we directly select the best hypothesis from the context-independent probability distribution using GPT. To run BERT-Defense without GPT, we select the hypothesis with the highest probability according to the results from the modified Levenshtein distance and improve it using context-dependent probabilities obtained with BERT. 
We evaluate on the rd:0.3,rd:0.3 attack scenario, 
because we think that it is the most challenging attack. 

The results are shown in Figure \ref{fig:ablation}.
They indicate that the most important component of BERT-Defense 
is the Levenshtein distance, as 
BERT 
often does not have enough context to meaningfully restore the sentences, given the difficult attacks from Zeroé that typically modify many words in each sentence. 
Removing BERT also considerably decreases the performance of the defense model. 
Finally, BERT-Defense without GPT performs on par with $\text{BD}_{agn}$ in these experiments, 
suggesting that BERT-Defense can also be used without GPT for hypothesis selection.

\paragraph{More illustrating examples.} 
To give an impression of the dataset and how the models cope with the adversarial attacks, 
we show more illustrating examples 
in Tables \ref{tab:morecases} and \ref{tab:othercases} (appendix). These indicate the superiority of our approach in that it typically generates semantically adequate sentences. 

%% file: conclusion.tex
\section{Conclusion}
We introduced \ours{}, a model that probabilistically combines context-independent word level information obtained from edit distance with context-dependent information from BERT's masked language modeling to combat low-level orthographic attacks. Our model does not train on possible error types but still substantially outperforms a spell-checker as well as the model of \citet{pruthi-etal-2019-combating}, which has been trained to shield against edit distance like attacks, on a comprehensive benchmark of cognitively inspired attack scenarios. We further show that our model rivals human crowd-workers supervised in a 3-shot manner. The generality of our approach allows it to be applied to a variety of different ``normalization'' problems, such as spelling normalization or OCR post-correction \citep{eger2016comparison} besides the adversarial attack scenario considered in this work, which we will explore in future work. 

We release our code and data at 
\url{https://github.com/yannikkellerde/BERT-Defense}.

%% file: appendix.tex
\appendix

\section{Appendices}
\subsection{Modified Wagner-Fischer algorithm}
\label{wagner-fischer}
The modified Wagner-Fischer algorithm gets the source word $S$ of length $n$ and the target word $T$ of length $m$ as inputs and performs the following operations in a run-time $O(mn)$.
\begin{enumerate}
    \item Initialize distance matrix 
    $\mathbf{D}$ of size $(m+1) \times (n+1)$
    with zeros
    \item For $i\in [1,m+1]$ do: $\mathbf{D_{i,1}}\gets i-1$
    \item Initialize a set-valued start matrix 
    $\mathbf{M}$ of the same size as $\mathbf{D}$ 
    with empty sets.
    \item For $j\in [1,n+1]$ do: $\mathbf{M_{1,j}}\gets \text{Set}\{j-1\}$
    \item For $i\in [1,m+1]$ and $j\in [1,n+1]$ do:
        \begin{itemize}
        \item Use previous entries of $D$ to calculate total cost of getting to $(i,j)$ with the edit distance operations:
            \begin{itemize}
                \item Insertion: $\mathbf{D_{i-1,j}}+1$
                \item Substitution: $\mathbf{D_{i-1,j-1}}+1$
                \item Deletion: $\mathbf{D_{i,j-1}}+1$
                \item Swap: $\mathbf{D_{i-2,j-2}}+1$
                \item If $T_i=S_j$ then no operation cost: $\mathbf{D_{i-1,j-1}}$
            \end{itemize}
        \item Enter the lowest cost from the edit distance operations into $\mathbf{D_{i,j}}$
        \item Update $\mathbf{M_{i,j}}$ by merging the set with the set-valued entries of $\mathbf{M}$ that led to $(i,j)$ with lowest cost
        \end{itemize}
    \item Initialize empty list $L$
    \item Store lowest entry of $\mathbf{D_{m+1}}$ as $c$ and for $j\in [1,n+1]$ do
    \begin{itemize}
        \item If $\mathbf{D_{m+1,j}=c}$ do: For $m\in\mathbf{M_{m+1,j}}$ do: Add a 2-tuple $(m,j)$ into L.
    \end{itemize}
    \item Return $c$,$L$
\end{enumerate}
\subsection{Visual and Phonetic attacks}
\label{app:vis_phon}
\paragraph{Visual attacks.} In the \ours{} full-distance pipeline, we exploit visual similarity (see appendix \ref{vis-sim}). The visual attacks implemented in \cite{benz} are also based on visual similarity. To avoid attacking with the same method that we defend with, we decided to use VIPER-DCES \cite{viper} instead. VIPER-DCES exchanges characters based on similarity of the descriptions from the Unicode 11.0.0 final names list (e.g. LATIN SMALL LETTER A for the character `a'). 
\paragraph{Phonetic attacks.} The phonetic embeddings implemented in \citet{benz} do not consistently produce phonetic attacks of sufficient quality. Thus, we used a many-to-many aligner \cite{m2m,eger2015we} together with the CMU Pronouncing Dictionary (cmudict) \cite{cmudict} and a word frequency list to calculate statistics for the correspondence between letters and phonemes. To attack a word, we convert the word to phonemes using cmudict and then convert it back to letters by sampling from the statistics. 
The perturbation probability $p$ for this attack controls the sampling temperature which describes how likely it is to sample letters that less frequently correspond to the phoneme in question. Using this method, we generate high-quality phonetically attacked sentences such as the one in Table \ref{tab:adv_attacks}.

\subsection{Visual similarity}
\label{vis-sim}
\begin{figure}[!t]
    \centering
    \includegraphics[width=1\columnwidth]{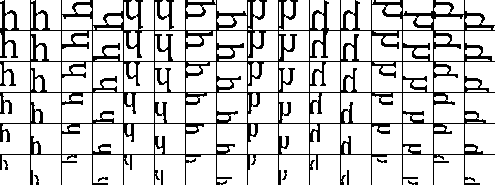}
    \caption{Different orientations/scales used for the letter h. The version that matches a Unicode character the best is used to calculate their similarity.}
    \label{fig:h_versions}
\end{figure}

We calculate the visual similarity of 30000 Unicode characters to 26 letters and 10 numbers. Each glyph is drawn with Python's pillow library \cite{pillow} in 20pt using a fitting font from the google-Noto font collection. The bitmap is then cropped to contain only the glyph. Then the image is resized and padded on the right and bottom to be of size $30px\times 30px$. When comparing the bitmap of a unicode glyph image and a letter/number glyph, multiple versions of the letter/number bitmap are created. For letters, the lowercase as well as the uppercase versions of each letter are taken. The bitmap gets downsized to 5 different sizes between $30px\times 30px$ and $15px\times 15px$, rotated and flipped in all 8 unique ways and then padded to $30px\times 30px$ again, such that the glyph is either placed at the top-left or the bottom left. See Figure \ref{fig:h_versions} for an example.
The percentage of matching black pixels between bitmaps are calculated and the highest matching percentage of all version becomes the similarity score $S$. The substitution cost between two characters will then be calculated based on the similarity with the equation $\textit{cost}=\max{(0,\min{(1,(0.8-S)*3)})}$. The parameters of this equation have been tuned, so that highly similar characters have a in very low substitution costs while weakly similar characters have next to no reduced in substitution cost.

\subsection{Parameters, runtime and computing infrastructure}
All experiments where run on a single machine using an  Intel(R) Core(TM) i7-4790K processor and a Nvidia GeForce GTX 1070 Ti graphics card. The restoration of a single sentence in the experiments took on average 0.1 seconds using ScRNN Defense, 1.34 seconds for Pyspellchecker and 8 seconds for BERT-defense. In total, $\text{BD}_{agn}$ includes 5 free parameters, most of them controlling the temperature of the used softmax operation to ensure good relative weighting of the probability distributions. The parameter values are shown in Table \ref{tab:parameters}. All additional parameters for $\text{BD}_{spec}$ have been described in \S\ref{sec:lev}. 
\begin{table}[!h]
\begin{tabularx}{\columnwidth}{l X}
 \toprule
 \textbf{Parameter} & \textbf{Value} \\ 
 \hline
 \makecell[l]{Softmax temperature for \\ context-independent hypothesis} & 10 \\
 \hline
 \makecell[l]{Softmax temperature for\\ context-independent word-probabilities} & 1 \\
 \hline
 Softmax temperature for BERT & 0.25 \\
 \hline
 Softmax temperature for GPT & 0.005 \\
 \hline
 Max number of hypothesis & 10\\
 \bottomrule
\end{tabularx}
\caption{Parameters for BERT-Defense.}
\label{tab:parameters}
\end{table}

\begin{table}[!h]
    \centering
    \noindent
\begin{tabularx}{\linewidth}{@{}>{\bfseries}p{2.8cm}@{\hspace{.5em}}X@{}}
\toprule
    Attacked & theensuing battls abd airstrikes killed at peast 10 militqnts.\\
    Ground-truth &  the ensuing battle and airstrikes killed at least 10 militants. \\ \midrule
    $\mathbf{BD_{agn}}$ & the ensuing battle and air strikes killed at least 10 militants. \\
    $\mathbf{BD_{spec}}$ & the ensuing battle and air strikes killed at least 10 militants. \\
    ScRNN Defense & tunney battls and airstrikes killed at past 10 militqnts. \\
    Pyspellchecker & theensuing battle abd airstrips killed at past 10 militants\\\bottomrule
    Attacked & \includegraphics[width=0.6\columnwidth,valign=b]{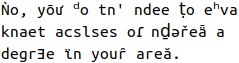}\\
    Ground-truth & No, you don't need to have taken classes or earned a degree in your area. \\ \midrule
    $\mathbf{BD_{agn}}$ & no, you do ,' nee ,' not besides of never a degree, you are. \\
    $\mathbf{BD_{spec}}$ & no, you do no' need to have taken classes or have a degree in your area. \\
    ScRNN Defense & , yu so to nerve to era knaet access of need a degree ïn your areă. \\
    Pyspellchecker & \includegraphics[width=0.6\columnwidth,valign=b]{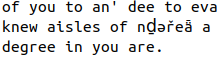}\\
    \bottomrule
    Attacked & A man ix riding ;n s voat.\\
    Ground-truth & A man is riding on a boat. \\ \midrule
    $\mathbf{BD_{agn}}$ & a man is riding in a boat. \\
    $\mathbf{BD_{spec}}$ & a man is riding in a boat. \\
    ScRNN Defense & a man imax riding on s voat. \\
    Pyspellchecker & a man ix riding in s vote\\
\bottomrule
\end{tabularx}
\caption{Additional adversarial shielding examples on the rd:0.3,rd:0.3 dataset.
}
\label{tab:othercases}
\end{table}

\begin{table*}[!t]
    \centering
    \noindent
\begin{tabularx}{\textwidth}{ |X||l|l|l|l|l|l|l|l|  }
 \hline
 \textbf{Dataset} &\multicolumn{2}{c}{$\mathbf{BD_{agn}}$}&\multicolumn{2}{|c}{$\mathbf{BD_{spec}}$}&\multicolumn{2}{|c}{\textbf{Pyspellchecker}}&\multicolumn{2}{|c|}{\textbf{ScRNN Defense}}\\
 \Xhline{2\arrayrulewidth}
 \textbf{Metric}&Mover&Editdist& Mover&Editdist&Mover&Editdist&Mover&Editdist\\\Xhline{2\arrayrulewidth}
 \textbf{vi:0.3}&0.696&5.04& 0.830&2.267&0.387&8.54&0.257&12.54\\\hline
 \textbf{tr:0.3}&0.778   &2.91&0.767 &3.14&0.605&3.34&0.386&7.25\\\hline
 \textbf{dv:0.3}&0.574   &9.27& 0.794&2.995&  0.335&9.53&0.379&9.48\\\hline
 \textbf{sg:0.3}&0.820   &2.02& 0.808&2.22&0.459&4.52&0.4&6.19\\\hline
 \textbf{is:0.3}& 0.539  &9.91& 0.842&2.767&0.44&9.26&0.520&7.04\\\hline
 \textbf{fs:0.3}& 0.399  &14.78& 0.688&3.227&0.310&14.41&0.277&16.12\\\hline
 \textbf{in:0.3}& 0.845  &1.9& 0.861&1.597&0.588&3.14&0.445&4.92\\\hline
 \textbf{kt:0.3}& 0.832  &1.96& 0.562&2.36&0.596&2.8&0.416&5.89\\\hline
 \textbf{nt:0.3}& 0.764  &3.38& 0.512&3.322&0.504&4.91&0.423&7.59\\\hline
 \textbf{ph:0.1}& 0.776  &3.21& 0.779&3.082&0.632&3.35&0.535&5.50\\\hline
 \textbf{ph:0.3}& 0.569  &8.77& 0.587&7.55&0.397&8.29&0.302&11.26\\\hline
 \textbf{ph:0.5}& 0.437  &13.25& 0.469&12.062&0.275&11.78&0.208&15.19\\\hline
 \textbf{ph:0.7}& 0.350  &16.37& 0.395&14.735&0.218&14.33&0.167&17.34\\\hline
 \textbf{ph:0.9}& 0.314  &18.45& 0.341&16.967&0.194&15.81&0.152&18.63\\\hline
 \textbf{sg:0.5,kt:0.3}& 0.701  &4.29& 0.537&4.47&0.23&7.07&0.158&10.25\\\hline
 \textbf{vi:0.3,in:0.3}& 0.627  &6.48& 0.679&2.467&0.172&13.73&0.14&15.75\\\hline
 \textbf{rd:0.3}& 0.676  &6.48& 0.650&3.485&0.44&7.25&0.375&8.91\\\hline
 \textbf{rd:0.3,rd:0.3}& 0.501  &11.49& 0.451&6.657&0.269&12.0425&0.232&13.92\\\hline
 \textbf{rd:0.6,rd:0.6}& 0.257  &22.30& 0.441&14.0&0.12&20.46&0.104&21.90\\
 \hline
\end{tabularx}
\caption{Exact scores for the results shown in Figure \ref{fig:evaluations_baseline}.}
\label{tab:exact_data}
\end{table*}